\documentclass[%
 reprint,
 amsmath,amssymb,
 aps,
]{revtex4-2}

\usepackage{graphicx, verbatim, multirow, enumitem,subfig}
\usepackage[usestackEOL]{stackengine}
\usepackage{dcolumn}
\usepackage{bm, url, hyperref}
\usepackage{times}

\begin{document}

\preprint{APS/123-QED}

\title{Limits of spectral learning under noise}

\author{Sabin Roman$^{1}$}
\author{Ljup{\v{c}}o Todorovski$^{2,1}$}
\author{Sa{\v{s}}o D{\v{z}}eroski$^{1}$}
\author{Marta Sales-Pardo$^{3,4}$}
\author{Roger Guimer\`a$^{3,4,5}$}
 \affiliation{$^{1}$Jo\u{z}ef Stefan Institute, Ljubljana, Slovenia}
 \affiliation{$^{2}$Faculty of Mathematics and Physics, University of Ljubljana, Slovenia}
 \affiliation{$^{3}$Department of Chemical Engineering, Universitat Rovira i Virgili,  Tarragona, Catalonia}
 \affiliation{$^{4}$Center for Computational Science and Applied Mathematics (ComSCIAM), Universitat Rovira i Virgili,  Tarragona, Catalonia}
 \affiliation{$^{5}$ICREA, Barcelona, Catalonia}


\begin{abstract}
Learning functional relationships from noisy data is a central problem in scientific inference. Spectral methods approximate unknown functions by expanding them in a basis and estimating the corresponding coefficients from data, but the stability of these coefficients under noise remains poorly understood. Here we study supervised regression with additive label noise using sparse spectral representations across multiple bases and dimensions. We show that noise induces a predictable drift in the learned coefficient vector whose magnitude depends on the effective number of active spectral modes. After whitening the empirical feature geometry, we derive a closed-form expression for the overlap between noisy and noiseless coefficient vectors, revealing a universal degradation curve governed by a single intrinsic noise scale. Numerical experiments across Fourier, Legendre, Bessel, and Haar bases confirm the theoretical prediction. The results demonstrate that spectral learning exhibits a fundamental noise threshold beyond which coefficient estimates become unstable, placing intrinsic limits on recovering functional structure from noisy data.
\end{abstract}

\maketitle


Interpretable mathematical models play a central role in our understanding of the world. With the advent of machine learning, developing methods for automatically learning such models from data has become an increasingly important direction in scientific research \cite{dzeroski07,evans10,cornelio23}, with applications ranging from physics \cite{schmidt2009distilling,cranmer20,reichardt2020bayesian,artime21} and chemistry \cite{minotaki24}, to engineering \cite{jog24} and even the study of complex social systems \cite{cabanas25}.

Among the variety of approaches developed to that end (including grammar-based \cite{brence2021probabilistic}, probabilistic \cite{guimera20,guimera26}, sparse regression \cite{brunton2016discovering} and deep learning-based \cite{udrescu2020ai,mevznar2023efficient} approaches, among others), a common strategy with well established roots in applied mathematics, physics and engineering is to approximate an unknown function using a spectral expansion in a chosen basis, yielding an interpretable decomposition into modes of increasing complexity \cite{roman2025approximating}. 
In practice, however, measurements are inevitably corrupted by noise, which perturbs the inferred spectral coefficients and can significantly distort the learned representation. 
Determining how noise propagates through spectral decompositions and limits the reliability of data-driven models is therefore a fundamental problem for scientific inference \cite{candes2006robust,belkin2019reconciling}.

Recent work has begun to investigate the limits of model discovery from noisy data, highlighting how measurement noise constrains the recovery of analytic structure from observations \cite{fajardo2023fundamental}. 
Theoretical studies of learning with noisy observations have also shown that noise can fundamentally limit the recoverability of underlying structures in data \cite{angluin1988learning,natarajan2013learning,omejc2024probabilistic}. 
Despite these advances, the mechanisms by which noise degrades spectral representations remain poorly understood. 
In particular, it is unclear how noise perturbs the geometry of the coefficient space and how this affects the stability of the inferred spectral decomposition.

Here, we study the degradation of spectral learning under additive noise, in a general situation in which: (i) training inputs \(x_i\in\Omega\subset\mathbb{R}^d\), \(i=1,\dots,N\), are sampled independently; (ii) the corresponding observed labels $y_i$ are given by
\begin{equation}
y_i=f(x_i)+\epsilon_i,
\qquad
\epsilon_i\sim\mathcal{N}(0,\sigma^2)\ \text{i.i.d.},
\label{eq:noise_model}
\end{equation}
with \(\sigma\ge 0\) controlling the noise amplitude; and (iii) we approximate $f(x)$ with a suitable spectral expansion 
on the basis functions \(\{\phi_k(x)\}_{k=1}^p\)
\begin{equation}
f_\sigma(x)
=
a(\sigma)+\sum_{k=1}^p c_k(\sigma)\,\phi_k(x),
\label{eq:model}
\end{equation}
where \(a(\sigma)\) is an intercept that is not included in the coefficient-space geometry, and \(\mathbf c(\sigma)\in\mathbb{R}^p\) contains the spectral coefficients.

Using such sparse spectral expansions, which have become central tools for identifying compact representations in high-dimensional feature spaces \cite{tibshirani1996regression,corywright24}, we analyze how the coefficient vector \(\mathbf c(\sigma)\) inferred from noisy data deviates from its noiseless counterpart. 
We show that, once the empirical feature geometry is appropriately transformed, the effect of noise can be interpreted as a stochastic drift in coefficient space. 
This perspective yields a universal analytic prediction for the overlap (or the distance) between noisy and noiseless coefficient vectors and identifies an intrinsic noise scale that governs the transition between stable and noise-dominated learning regimes.

\medskip

\begin{figure*}[t!]
    \centering 
    \subfloat[1D example: $x^2$ with a Fourier expansion]{
        \includegraphics[width=0.47\textwidth]{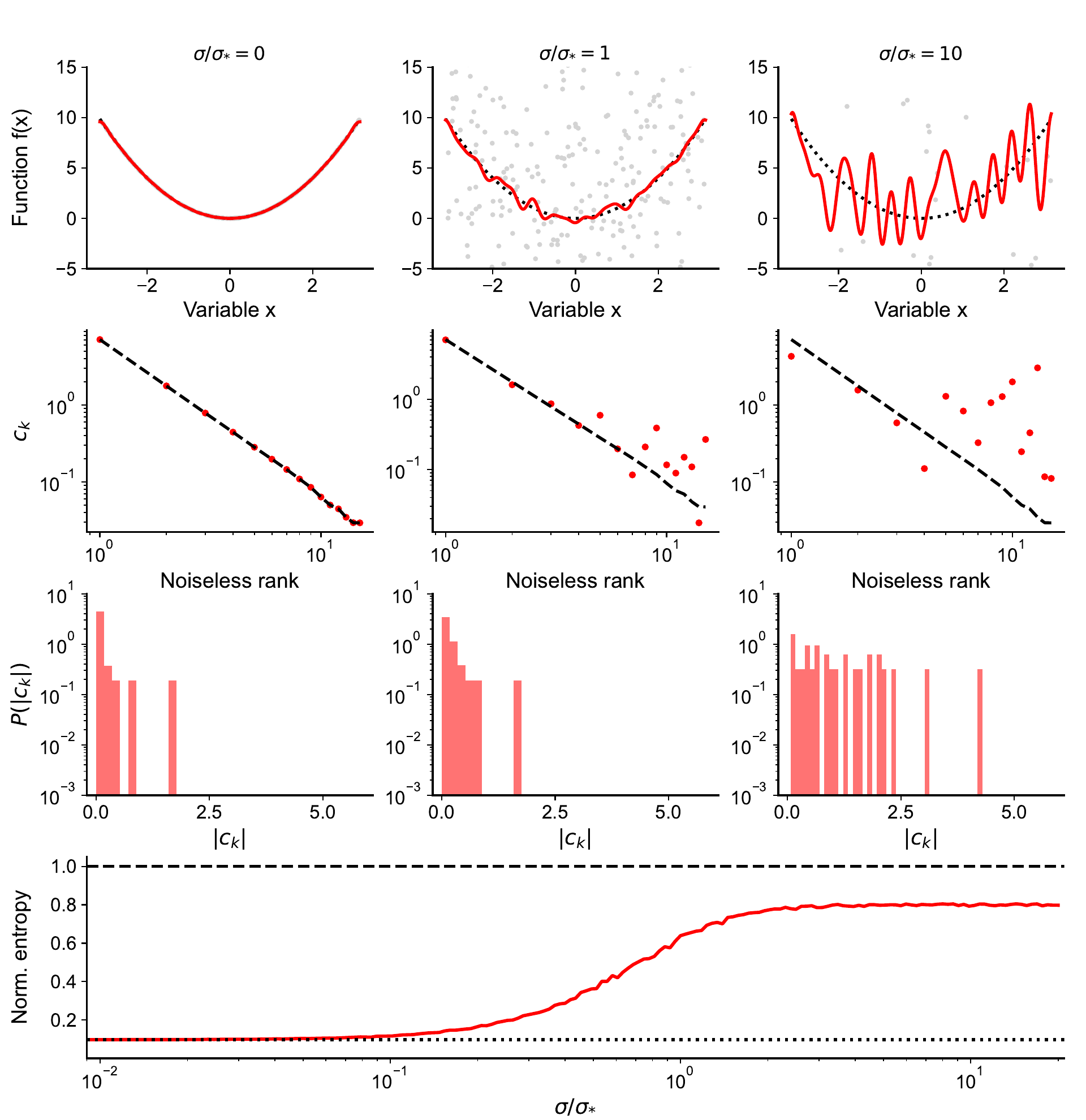}
    }
    \hfill
    \subfloat[2D example: $\sin(x_{1}^2e^{x_{2}})$ expanded in a Legendre basis]{
        \includegraphics[width=0.47\textwidth]{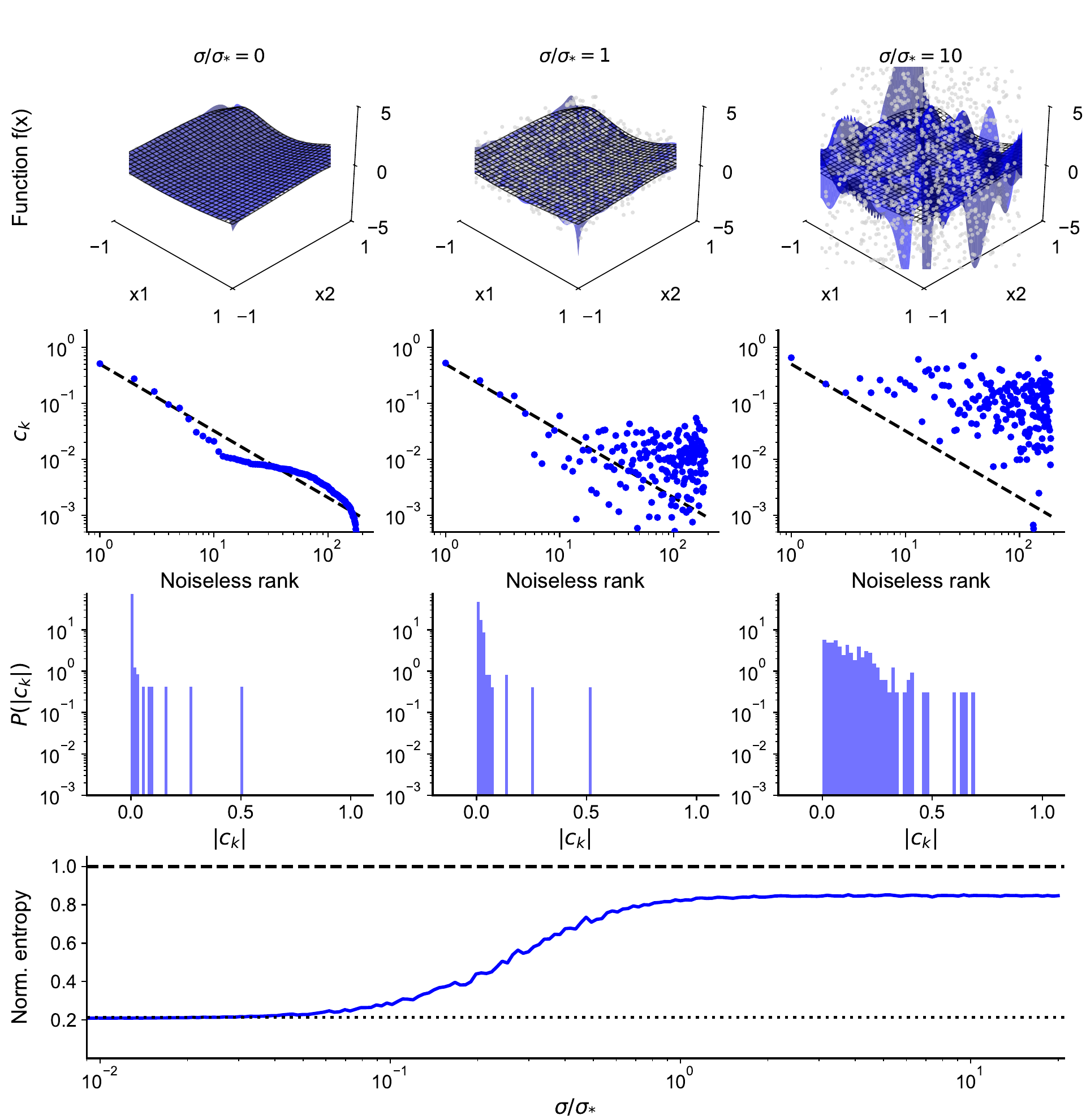}
    }

    \caption{
Examples of spectral learning under increasing noise for: (a) one-dimensional; (b) two-dimensional target functions. 
The top rows show the true function (dotted) and a representative reconstructed function learned from noisy data (grey dots) for increasing relative noise levels $\sigma/\sigma^*$, where $\sigma^*$ is the characteristic noise scale derived in Eq.~(\ref{eq:sigma_star}). 
The second and third rows summarize the behavior of the spectral coefficients across multiple independent noise realizations: the second row shows the median coefficient magnitudes $|c_k|$ as a function of coefficient index, while the third row shows the corresponding median distribution of coefficient magnitudes. 
The bottom row displays the normalized spectral entropy of the power spectrum $|c_k|^2$ as a function of noise, computed as the median over noise realizations. 
Horizontal reference lines indicate theoretical bounds corresponding to a sparse power-law spectrum (lower bound) and to a uniform distribution of coefficients (upper bound). As the noise level increases, the learned spectrum broadens and spectral weight spreads across modes, leading to an increase in entropy and a progressive loss of spectral structure.
    }
    \label{fig:1}
\end{figure*}
To illustrate how noise affects spectral learning, we first examine sparse spectral expansions as applied to simple one- and two-dimensional regression problems, where we use either Fourier or orthonormal Legendre bases adapted to the structure and symmetries of the target function.
In Fig.~\ref{fig:1}, we show the learned functions, coefficient spectra, and coefficient distributions for increasing levels of relative noise $\sigma/\sigma_*$. 
Here $\sigma_*$ denotes the characteristic noise scale at which the noise-induced coefficient perturbation becomes comparable to the noiseless spectral signal; we derive it explicitly below. 
In the noiseless case, the learned function accurately reproduces the target, and the inferred spectral coefficients follow the expected power-law decay (Fig.~\ref{fig:1}, first and second rows) \cite{stein2003fourier}. 
As noise increases, the inferred functions become progressively distorted and the spectral coefficients deviate strongly from the noiseless scaling. 
These effects appear already for moderate noise levels and become pronounced once the noise amplitude approaches the characteristic scale $\sigma_*$.

To characterize the typical behavior across noise realizations, the coefficient statistics shown in Fig.~\ref{fig:1} are computed using median values over multiple independent noise draws. 
The median coefficient spectrum and the associated coefficient distributions reveal that noise does not simply introduce random fluctuations, but instead induces a systematic redistribution of spectral weight. 
Indeed, in both one and two dimensions, the initially sparse power-law spectrum gradually broadens as noise increases (Fig.~\ref{fig:1}, second and third rows). 
This redistribution is captured by the normalized entropy of the coefficient power spectrum, which increases sharply with noise (Fig.~\ref{fig:1}, bottom row). 
Indeed, at low noise levels the entropy remains close to the lower bound associated with a sparse power-law spectrum, while for large noise it approaches the value expected for a 
uniform distribution of coefficients.

We aim to understand this degradation of the spectral expansion. To that end, we start by recalling why sparse regression leads to consistent recovery of the spectral coefficients. 
As mentioned above, we consider supervised regression with additive label noise as described in Eq.~\eqref{eq:noise_model}, and approximate the target function $f(x)$ using the spectral expansion in Eq.~\eqref{eq:model}.
%
The design matrix associated to this spectral expansion contains only the nonconstant basis functions $\Phi_{ik}=\phi_k(x_i),
\Phi\in\mathbb{R}^{N\times p}$.
We refer to the columns of \(\Phi\), namely the sampled basis functions \((\phi_k(x_1),\ldots,\phi_k(x_N))^\top\), as the empirical features; we center the nonconstant empirical features by defining
\[
\boldsymbol\mu=\frac1N\Phi^\top\mathbf 1
\qquad\text{and}\qquad
\Phi_c=\Phi-\mathbf 1\boldsymbol\mu^\top .
\]
The labels are centered analogously, and the intercept $a(\sigma)$ is recovered separately.
 

From the centered empirical features, we further define the empirical Gram matrix $G_c=\Phi_c^\top\Phi_c/N$, which determines the natural geometry of spectral coefficient space, because the mean squared error between two spectral expansions with coefficients $\mathbf{c}$ and $\mathbf{c'}$ is \footnote{Here and throughout the manuscript, the norm $\|\dots\|$ without subindices indicates the usual L2 norm.}
\[
\frac{1}{N}\|\Phi_c \mathbf c- \Phi_c \mathbf c'\|^2
=
(\mathbf c-\mathbf c')^\top G_c(\mathbf c-\mathbf c'),
\]
so differences between coefficient vectors are weighted by $G_c$. 


If the centered empirical features were exactly orthonormal, then \(G_c=I\). 
This idealized limit explains why sparse regression leads to consistent spectral learning---in this situation, the coordinates decouple and, under the noise-correlation condition
\[
\left\|\frac1N\Phi_c^\top \boldsymbol\epsilon\right\|_\infty \le \Delta,
\]
the Lasso regression solution satisfies \cite{hastie2015statistical}
\begin{equation*}
\|\hat{\mathbf c}(\sigma)-\mathbf c_0\|_\infty \le 2\Delta,
\label{eq:l_inf_bound}
\end{equation*}
or, equivalently, \( |\hat c_k(\sigma)-(\mathbf c_0)_k|\le 2\Delta \) for all \(k\), where \(\mathbf c_0 := \hat{\mathbf c}(0)\) denotes the noiseless reference coefficient vector.

At finite sample size, however, and even though the chosen bases are orthonormal in the population inner product, \(G_c\) generally differs from the identity because sampling fluctuations induce correlations between empirical features. 
The issue is therefore how noise alters the coefficient vector when the empirical feature geometry is determined by \(G_c\), rather than by the standard Euclidean norm. 

To address this question, we restore an approximately Euclidean geometry by whitening the centered empirical feature covariance \cite{kessy2018optimal}---we introduce
\begin{equation*}
A=(G_c+\rho I)^{-1/2},
\label{eq:A_def}
\end{equation*}
where $\rho>0$ is a small ridge parameter introduced for numerical stability, and define the whitened features and coefficients
\begin{equation*}
\tilde\Phi=\Phi_c A,
\qquad
\mathbf c=A\tilde{\mathbf c}.
\label{eq:transform}
\end{equation*}
In these new coordinates, the empirical Gram matrix becomes
\begin{equation*}
\tilde G
=
\frac{1}{N}\tilde\Phi^\top\tilde\Phi
=
A^\top G_c A
\approx I .
\label{eq:whitened_gram}
\end{equation*}
Thus, the approximate identity that cannot be assumed for the original finite-sample design is now enforced by whitening the centered empirical geometry.

We solve the sparse regression on the centered-whitened design by finding
\begin{equation*}
\hat{\tilde{\mathbf c}}(\sigma) =
\arg\min_{\tilde{\mathbf c}\in\mathbb{R}^p}
\left[
\frac{1}{2N}\|\tilde\Phi\tilde{\mathbf c}-\mathbf y_c\|^2
+\lambda\|\tilde{\mathbf c}\|_1
\right],
\label{eq:lasso}
\end{equation*}
where \(\mathbf y_c\) denotes the centered label vector. 
Throughout the noise sweep, the regularization parameter \(\lambda\) is held fixed, so that changes in the recovered coefficients reflect the effect of label noise on a fixed sparse spectral estimator rather than changes in model complexity induced by retuning the estimator at each noise level. 
All geometric quantities below, including overlaps, coefficient distances, \(p_{\mathrm{eff}}\), and \(\sigma_*\), are computed in the centered-whitened coordinates.

This whitening makes it possible to describe the effect of label noise as a perturbation problem in coefficient space. 
Writing the noisy solution as \(\tilde{\mathbf c}(\sigma)=\tilde{\mathbf c}_0+\delta\tilde{\mathbf c}(\sigma)\), where \(\tilde{\mathbf c}_0:=\hat{\tilde{\mathbf c}}(0)\), the noise enters through the forcing term \((1/N)\tilde\Phi^\top \boldsymbol\epsilon\), whose covariance is
\begin{equation*}
\mathrm{Cov}\!\left[
\frac{1}{N}\tilde\Phi^\top \boldsymbol\epsilon
\right]
=
\frac{\sigma^2}{N^2}\tilde\Phi^\top\tilde\Phi
=
\frac{\sigma^2}{N}\tilde G
\approx
\frac{\sigma^2}{N}I.
\label{eq:noise_cov}
\end{equation*}
Indeed, since $\tilde{G}\approx I$, the normal equations in whitened coordinates reduce to approximately $\delta\tilde{\mathbf{c}}\approx \frac{1}{N}\tilde\Phi^\top\boldsymbol\epsilon$, so each component $(\delta\tilde{\mathbf{c}})_k$ is a zero-mean random variable with variance $\sigma^2/N$, and components across distinct directions are approximately uncorrelated.
Since $\boldsymbol\epsilon$ is mean-zero, so is $\delta\tilde{\mathbf{c}}$, and the expected squared norm expands as
\begin{equation}
\mathbb{E}\|\delta\tilde{\mathbf c}\|^2
\approx
\sum_{k=1}^{p_{\mathrm{eff}}} \frac{\sigma^2}{N}
=
\frac{\sigma^2}{N}\,p_{\mathrm{eff}},
\label{eq:delta_norm}
\end{equation}
where \(p_{\mathrm{eff}} = \big|\{k : |(\tilde{\mathbf c}_0)_k| > \tau\}\big|\) is the number of significantly excited directions in the noiseless solution 
and the sum is restricted to active directions since inactive ones contribute negligibly.
Equivalently, \(\|\delta\tilde{\mathbf c}\|\sim \sigma\sqrt{p_{\mathrm{eff}}/N}\).
To find the intrinsic noise scale, we compare this perturbation magnitude with the signal norm \(\|\tilde{\mathbf c}_0\|\). 
So that, by defining
\[
r(\sigma)
=
\frac{\sqrt{\mathbb{E}\|\delta\tilde{\mathbf c}\|^2}}{\|\tilde{\mathbf c}_0\|},
\]
and using Eq.~\eqref{eq:delta_norm}, we obtain
\begin{equation}
r(\sigma)
\approx
\frac{\sigma}{\sigma_*},
\qquad
\sigma_*
=
\|\tilde{\mathbf c}_0\|
\sqrt{\frac{N}{p_{\mathrm{eff}}}},
\label{eq:sigma_star}
\end{equation}
where \(\sigma_*\) is the noise amplitude at which the perturbation norm becomes comparable to the signal norm in centered-whitened coefficient space.

\begin{figure*}[t]
  \centering

  \subfloat[\label{fig:overlap_abs}Overlap $q$ vs.\ $\sigma$]{
    \includegraphics[width=0.42\textwidth]{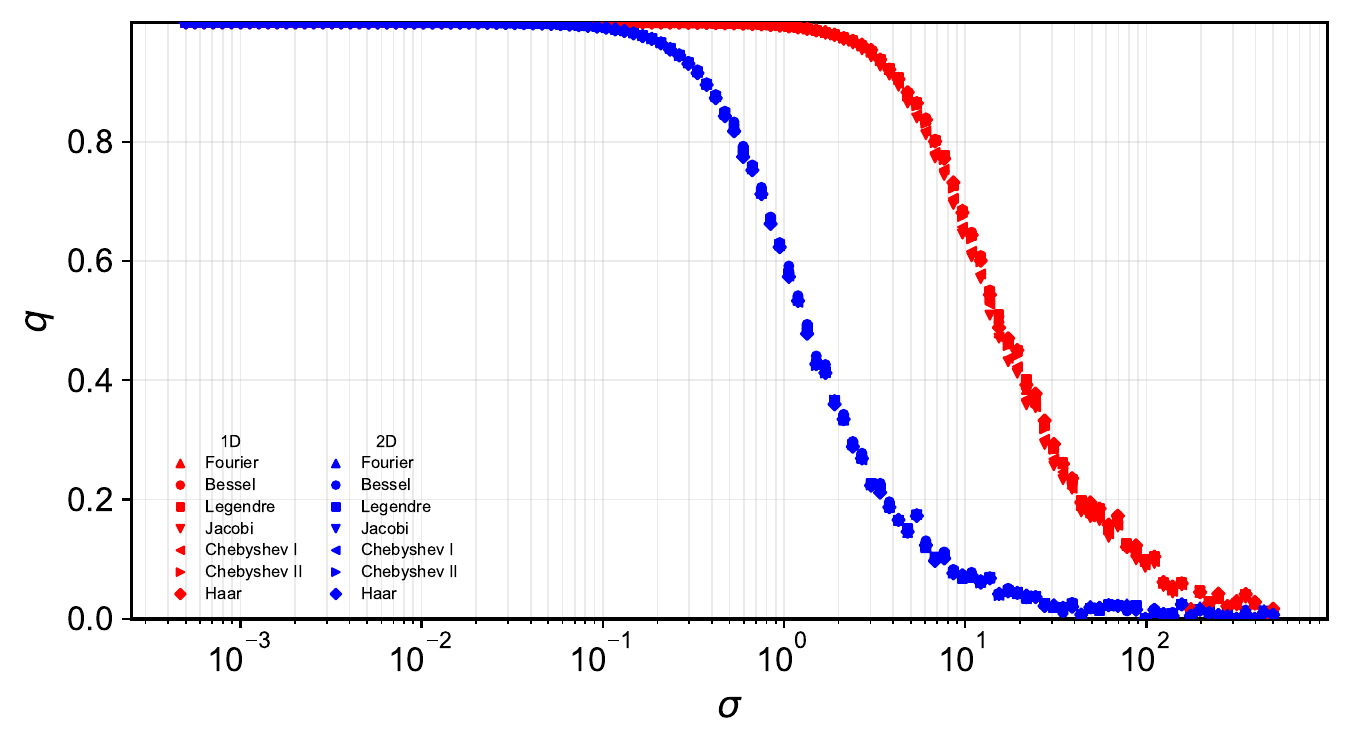}
  }
  \subfloat[\label{fig:overlap_rel}Overlap $q$ vs.\ $\sigma/\sigma^*$]{
    \includegraphics[width=0.42\textwidth]{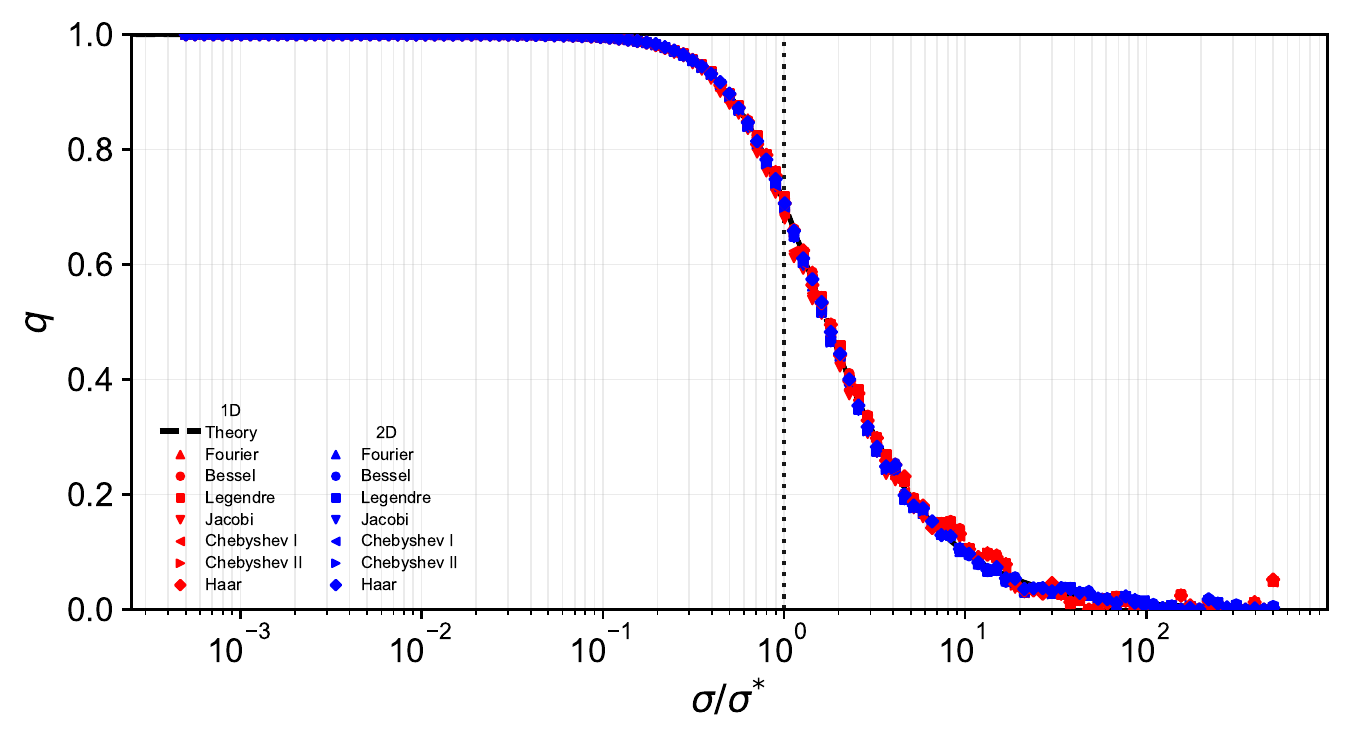}
  }\\[-0.5ex]

  \subfloat[\label{fig:rmse_abs}RMSE vs.\ $\sigma$]{
    \includegraphics[width=0.42\textwidth]{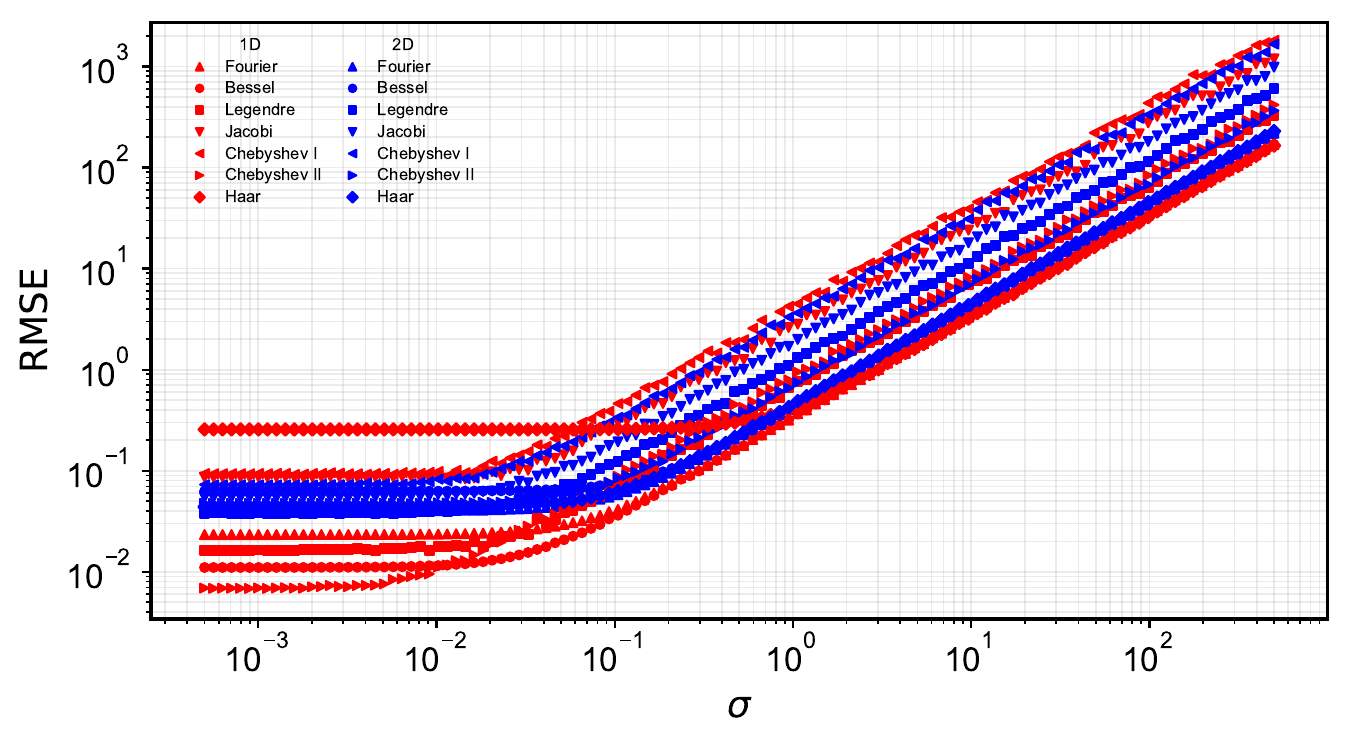}
  }
  \subfloat[\label{fig:dist_rel}Coeff.\ distance vs.\ $\sigma/\sigma^*$]{
    \includegraphics[width=0.42\textwidth]{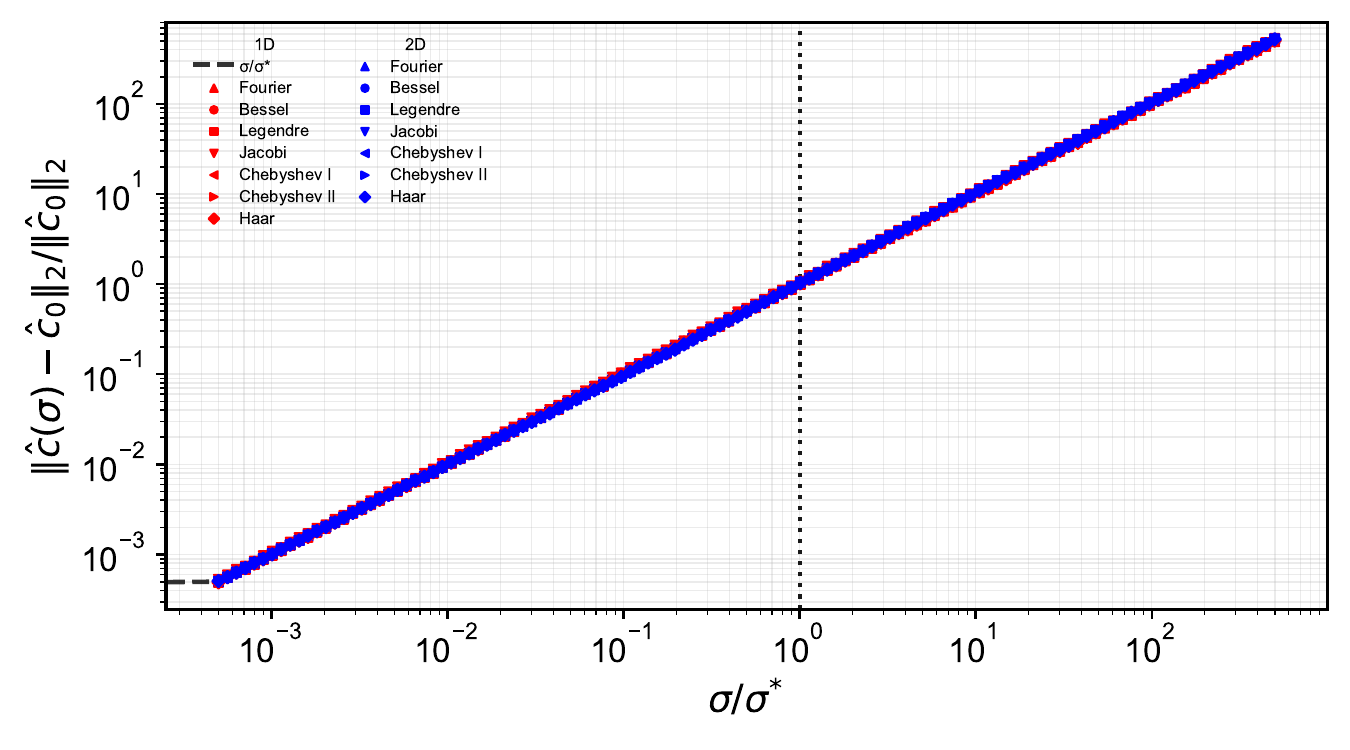}
  }

\caption{Spectral degradation under noise across bases and dimensions. 
We show results for bases on compact support: Fourier, Bessel, Legendre, Jacobi, Chebyshev (I and II) and Haar bases in one and two dimensions.
(a,b) Cosine overlap $q$ between the noisy and noiseless centered-whitened coefficient vectors. 
When expressed in terms of $\sigma/\sigma^*$, the empirical curves collapse across bases 
and dimensions onto the theoretical prediction in Eq.~\eqref{eq:q_theory}.
(c) Test root-mean-square error (RMSE) as a function of the noise amplitude $\sigma$. 
(d) Normalized Euclidean distance between the noisy and noiseless centered-whitened coefficient vectors, 
$\|\tilde{\mathbf c}(\sigma)-\tilde{\mathbf c}_0\|_2/\|\tilde{\mathbf c}_0\|_2$, as a function of $\sigma/\sigma^*$. 
Across bases and dimensions, both RMSE and coefficient distance increase consistently with the predicted noise-induced drift in coefficient space governed by the intrinsic noise scale $\sigma^*$.}
  \label{fig:2}
\end{figure*}

To quantify stability, we consider the (cosine) overlap  $q(\sigma)$ between noisy and noiseless centered-whitened coefficient vectors,
\begin{equation}
q(\sigma)
=
\frac{\tilde{\mathbf c}(\sigma)\cdot\tilde{\mathbf c}_0}
{\|\tilde{\mathbf c}(\sigma)\|\,\|\tilde{\mathbf c}_0\|}.
\label{eq:q_def}
\end{equation}
Assuming that \(\delta\tilde{\mathbf c}\) is 
approximately isotropic in the active subspace, the cross term averages to zero and \(\|\tilde{\mathbf c}(\sigma)\|^2\approx \|\tilde{\mathbf c}_0\|^2+\|\delta\tilde{\mathbf c}\|^2\). 
Then, Eq.~\eqref{eq:q_def} gives 
\begin{equation}
q(\sigma)\approx \left(
1+
r(\sigma)^2
\right)^{-1/2} = 
\left(
1+
\left(\frac{\sigma}{\sigma_*}\right)^2
\right)^{-1/2}.
\label{eq:q_theory}
\end{equation}
This expression contains no adjustable parameters once \(\tilde{\mathbf c}_0\), \(p_{\mathrm{eff}}\), and \(N\) are fixed, and matches empirical results across different bases and dimensions (Fig.~\ref{fig:overlap_rel}).

The same perturbative model predicts the normalized coefficient drift,
\begin{equation}
\frac{\|\tilde{\mathbf c}(\sigma)-\tilde{\mathbf c}_0\|}
{\|\tilde{\mathbf c}_0\|}
\sim
\frac{\sigma}{\sigma_*},
\label{eq:distance_scaling_normalized}
\end{equation}
which we observe empirically across bases and dimensions (Fig.~\ref{fig:dist_rel}). 
In centered-whitened coordinates the quadratic loss satisfies
\begin{equation*}
\frac{1}{N}\|\tilde\Phi(\tilde{\mathbf c}-\tilde{\mathbf c}_0)\|^2
=
(\tilde{\mathbf c}-\tilde{\mathbf c}_0)^\top
\tilde G
(\tilde{\mathbf c}-\tilde{\mathbf c}_0)
\approx
\|\tilde{\mathbf c}-\tilde{\mathbf c}_0\|^2,
\label{eq:loss_distance}
\end{equation*}
so the noise-induced contribution to the prediction error follows the same scaling as
the coefficient drift. The test RMSE shown in Fig.~\ref{fig:rmse_abs} includes both this noise-induced
term and a baseline approximation error that is present even at $\sigma=0$, arising from
the finite basis truncation and the regularization bias of the fixed-$\lambda$ estimator. At low noise the baseline dominates and the RMSE saturates at a floor, whereas the normalized coefficient distance $\|\tilde{\mathbf c}(\sigma)-\tilde{\mathbf c}_0\|/\|\tilde{\mathbf c}_0\|$
has no such floor because it is defined relative to $\tilde{\mathbf c}_0$ and is exactly zero at $\sigma=0$ by construction. Once the noise-induced term becomes comparable to the baseline, the RMSE begins to grow and follows the same $\sigma/\sigma_*$ scaling
(Fig.~\ref{fig:rmse_abs}).

Fig.~2(a,b) shows the empirical overlap \(q(\sigma)\) obtained from sparse spectral regression using Fourier, Bessel, Legendre, Jacobi, Chebyshev (I and II) and Haar bases in one and two dimensions. 
When plotted against \(\sigma\) (Fig.~\ref{fig:overlap_abs}), the overlap varies across bases and dimensions. 
When plotted against \(\sigma/\sigma_*\) (Fig.~\ref{fig:overlap_rel}), the curves collapse onto Eq.~\eqref{eq:q_theory}, indicating that \(\sigma_*\) captures the dominant dependence of spectral degradation on the noiseless coefficient geometry. 
Fig.~\ref{fig:dist_rel} shows the corresponding normalized coefficient distance, which grows linearly with \(\sigma/\sigma_*\), while Fig.~\ref{fig:rmse_abs} shows the associated growth of the test RMSE.

The results above provide a geometric interpretation of noise-induced degradation in spectral learning. 
Label noise induces an approximately isotropic perturbation in centered-whitened spectral coordinates, whose magnitude is controlled by the intrinsic noise scale $\sigma_\ast$ defined in Eq.~\eqref{eq:sigma_star}. 
When $\sigma \ll \sigma_\ast$, the perturbation remains small compared with the signal and the learned spectrum is stable; when $\sigma \gtrsim \sigma_\ast$, noise dominates the coefficient dynamics and the spectral structure progressively deteriorates. 
Because this derivation relies only on the centered empirical feature geometry and on the propagation of label noise through the regression problem, the resulting scaling laws apply broadly to spectral learning methods based on orthogonal function expansions.

These findings identify a fundamental noise scale that governs the stability of spectral representations learned from data. 
Beyond this scale, recovering detailed spectral structure becomes intrinsically difficult, as noise redistributes spectral weight across modes and increases the effective dimensionality of the representation. 

It is instructive to contrast this behavior with symbolic regression approaches, where one aims to identify closed-form models from data, rather than approximating them via spectral expansions.
Despite addressing different learning problems, the present spectral
analysis and the symbolic-regression results of Fajardo-Fontiveros \textit{et al.}~\cite{fajardo2023fundamental} identify characteristic noise scales with the same basic structure
\[
\small
\text{noise scale} \;\sim\; \text{intrinsic data scale}\,\sqrt{\frac{\text{number of data points }}{\text{effective model dimension}}}.
\]
In our case the intrinsic scale is set by the norm of the noiseless centered-whitened coefficient vector, while in symbolic regression it is naturally related to the variation of the true function $f(x)$. 
The square-root factor captures the same balance between data and complexity: increasing the number of data points raises the tolerable noise level as $\sqrt{N}$, whereas increasing the effective model dimension lowers it through the square root of the number of active modes. 
Both approaches therefore point to the same trade-off between intrinsic signal scale, sample size, and model complexity.

The two settings differ, however, in the geometry of the representation space. 
Closed-form expressions generated in symbolic regression frameworks typically involve nonzero terms at every order when expanded in a generic functional basis, whereas symmetry-adapted orthogonal expansions naturally exhibit structured sparsity, for example through coefficients that vanish by symmetry. 
If the signs of expansion coefficients are encoded as digits in base three, mapping negative, zero, and positive coefficients to $\{0,1,2\}$, symbolic-regression candidates correspond to points in a sparse, fractal-like subset of coefficient space analogous to the Cantor set, while structured spectral expansions populate a denser set of admissible coefficient patterns. 
This suggests that the abrupt noise-driven transition reported by~\cite{fajardo2023fundamental} may reflect the gapped topology of symbolic representation space, whereas the orthogonal expansions studied here exhibit continuous degradation with noise. 
From this perspective, it seems important to study in more depth how the topology of the representation space of mathematical models may help determine how noise limits the recovery of functional structure from data.

\textbf{Code availability:} \url{https://zenodo.org/records/20637540}.

\section*{Disclaimer}
Co-funded by the European Union. Views and opinions expressed are however those of the author(s) only and do not necessarily reflect those of the European Union or European Research Executive Agency. Neither the European Union nor the granting authority can be held responsible for them.

\section*{Funding}
This work is supported by EU's Horizon Europe under the Marie Sk\l{}odowska-Curie Postdoctoral Fellowship Programme, SMASH co-funded under grant agreement No.~101081355. The operation (SMASH project) is co-funded by the Republic of Slovenia and the European Union from the European Regional Development Fund. LT and SD acknowledge support by the Slovenian Research Agency via the research program Knowledge Technologies (P2-0103) and the Gravity project AI for Science (GC-0001). MSP and RG acknowledge support by project PID2022-142600NB-I00 from MCIN/ AEI/ 10.13039/ 501100011033, and by the Government of Catalonia (2021SGR-633).

\bibliographystyle{apsrev4-2}
\bibliography{references}

@PREAMBLE{
 "\providecommand{\noopsort}[1]{}" 
 # "\providecommand{\singleletter}[1]{#1}%" 
}

@book{stein2003fourier,
  author    = {Stein, Elias M. and Shakarchi, Rami},
  title     = {Fourier Analysis: An Introduction},
  series    = {Princeton Lectures in Analysis},
  volume    = {1},
  publisher = {Princeton University Press},
  year      = {2003}
}

@article{fajardo2023fundamental,
  title={Fundamental limits to learning closed-form mathematical models from data},
  author={Fajardo-Fontiveros, Oscar and Reichardt, Ignasi and De Los R{\'\i}os, Harry R and Duch, Jordi and Sales-Pardo, Marta and Guimer{\`a}, Roger},
  journal={Nat. Comm.},
  volume={14},
  number={1},
  pages={1043},
  year={2023},
  publisher={Nature Publishing Group UK London},
  doi={http://dx.doi.org/10.1038/s41467-023-36657-z}
}

@article{kessy2018optimal,
  title={Optimal whitening and decorrelation},
  author={Kessy, Agnan and Lewin, Alex and Strimmer, Korbinian},
  journal={Am. Stat.},
  volume={72},
  number={4},
  pages={309--314},
  year={2018},
  publisher={Taylor \& Francis},
  doi={https://doi.org/10.1080/00031305.2016.1277159}
}

@article{schmidt2009distilling,
  title={Distilling free-form natural laws from experimental data},
  author={Schmidt, Michael and Lipson, Hod},
  journal={Science},
  volume={324},
  number={5923},
  pages={81--85},
  year={2009},
  doi={https://doi.org/10.1126/science.1165893}
}

@article{udrescu2020ai,
  title={AI Feynman: A physics-inspired method for symbolic regression},
  author={Udrescu, Silviu-Marian and Tegmark, Max},
  journal={Sci. Adv.},
  volume={6},
  number={16},
  pages={eaay2631},
  year={2020},
  doi={https://doi.org/10.1126/sciadv.aay2631}
}

@article{belkin2019reconciling,
  title={Reconciling modern machine-learning practice and the classical bias--variance trade-off},
  author={Belkin, Mikhail and Hsu, Daniel and Ma, Siyuan and Mandal, Soumik},
  journal={Proc. Natl. Acad. Sci. USA},
  volume={116},
  number={32},
  pages={15849--15854},
  year={2019},
  publisher={National Academy of Sciences},
  doi={https://doi.org/10.1073/pnas.1903070116}
}

@article{candes2006robust,
  title={Robust uncertainty principles: Exact signal reconstruction from highly incomplete frequency information},
  author={Cand{\`e}s, Emmanuel J and Romberg, Justin and Tao, Terence},
  journal={IEEE Trans. Inf. Theory},
  volume={52},
  number={2},
  pages={489--509},
  year={2006},
  publisher={IEEE},
  doi={https://doi.org/10.1109/TIT.2005.862083}
}

@article{tibshirani1996regression,
  title={Regression shrinkage and selection via the {L}asso},
  author={Tibshirani, Robert},
  journal={J. R. Stat. Soc. Ser. B Stat. Methodol.},
  volume={58},
  number={1},
  pages={267--288},
  year={1996},
  publisher={Oxford University Press},
  doi={https://doi.org/10.1111/j.2517-6161.1996.tb02080.x}
}

@book{hastie2015statistical,
  title={Statistical Learning with Sparsity: The Lasso and Generalizations},
  author={Hastie, Trevor and Tibshirani, Robert and Wainwright, Martin},
  year={2015},
  publisher={CRC Press},
  address={Boca Raton, FL}
}

@article{brunton2016discovering,
  title={Discovering governing equations from data by sparse identification of nonlinear dynamical systems},
  author={Brunton, Steven L and Proctor, Joshua L and Kutz, J Nathan},
  journal={Proc. Natl. Acad. Sci. USA},
  volume={113},
  number={15},
  pages={3932--3937},
  year={2016},
  publisher={National Academy of Sciences},
  doi={https://doi.org/10.1073/pnas.1517384113}
}

@article{natarajan2013learning,
  title={Learning with noisy labels},
  author={Natarajan, Nagarajan and Dhillon, Inderjit S and Ravikumar, Pradeep K and Tewari, Ambuj},
  journal={Adv. Neural Inf. Process. Syst.},
  volume={26},
  year={2013}
}

@article{angluin1988learning,
  title={Learning from noisy examples},
  author={Angluin, Dana and Laird, Philip},
  journal={Mach. Learn.},
  volume={2},
  number={4},
  pages={343--370},
  year={1988},
  publisher={Springer},
  doi={https://doi.org/10.1007/BF00116829}
}

@article{roman2025approximating,
  title={Approximating the universal thermal climate index using sparse regression with orthogonal polynomials},
  author={Roman, Sabin and Skok, Gregor and Todorovski, Ljupco and Dzeroski, Saso},
  journal={arXiv preprint arXiv:2508.11307},
  year={2025}
}

@article{reichardt2020bayesian,
title = {{B}ayesian machine scientist to compare data collapses for the {N}ikuradse dataset},
author = {Reichardt, I. and Pallar\`es, J. and Sales-Pardo, M. and Guimer\`a, R.},
journal = {Phys. Rev. Lett.},
volume = {124},
pages = {084503},
year = {2020},
doi = {http://dx.doi.org/10.1103/PhysRevLett.124.084503}
}

@article{brence2021probabilistic,
  title={Probabilistic grammars for equation discovery},
  author={Brence, Jure and Todorovski, Ljup{\v{c}}o and D{\v{z}}eroski, Sa{\v{s}}o},
  journal={Knowl.-Based Syst.},
  volume={224},
  pages={107077},
  year={2021},
  publisher={Elsevier},
  doi={https://doi.org/10.1016/j.knosys.2021.107077}
}

@article{omejc2024probabilistic,
  title={Probabilistic grammars for modeling dynamical systems from coarse, noisy, and partial data},
  author={Omejc, Nina and Gec, Bo{\v{s}}tjan and Brence, Jure and Todorovski, Ljup{\v{c}}o and D{\v{z}}eroski, Sa{\v{s}}o},
  journal={Mach. Learn.},
  volume={113},
  number={10},
  pages={7689--7721},
  year={2024},
  publisher={Springer},
  doi={https://doi.org/10.1007/s10994-024-06522-1}
}

@article{mevznar2023efficient,
  title={Efficient generator of mathematical expressions for symbolic regression},
  author={Me{\v{z}}nar, Sebastian and D{\v{z}}eroski, Sa{\v{s}}o and Todorovski, Ljup{\v{c}}o},
  journal={Mach. Learn.},
  volume={112},
  number={11},
  pages={4563--4596},
  year={2023},
  publisher={Springer},
  doi={https://doi.org/10.1007/s10994-023-06400-2}
}

@Book{dzeroski07,
  editor = 	 {S. D\v{z}eroski and L. Todorovski},
  title = 	 {Computational Discovery of Scientific Knowledge},
  publisher = 	 {Springer},
  year = 	 {2007},
  series = 	 {Lecture Notes in Artificial Intelligence},
}

@article{guimera20,
    title = {A {B}ayesian machine scientist to aid in the solution of challenging scientific problems},
    author = {Guimer\`a, R. and Reichardt, I. and Aguilar-Mogas, A. and Massucci, F. A. and Miranda, M. and Pallar\`es, J. and Sales-Pardo, M.},
    journal = {Sci. Adv.},
    volume = {6},
    pages = {eaav6971},
    year = {2020},
    doi = {http://dx.doi.org/10.1126/sciadv.aav6971}
}

@article{evans10,
    author = {Evans, James and Rzhetsky, Andrey},
    journal = {Science},
    number = {5990},
    pages = {399--400},
    pmid = {20651141},
    title = {Machine science},
    volume = {329},
    year = {2010},
    doi={https://doi.org/10.1126/science.1189416}
}

@inproceedings{cranmer20,
author = {Cranmer, Miles and Sanchez-Gonzalez, Alvaro and Battaglia, Peter and Xu, Rui and Cranmer, Kyle and Spergel, David and Ho, Shirley},
title = {Discovering symbolic models from deep learning with inductive biases},
year = {2020},
isbn = {9781713829546},
publisher = {Curran Associates Inc.},
address = {Red Hook, NY, USA},
booktitle = {Proceedings of the 34th International Conference on Neural Information Processing Systems},
articleno = {1462},
numpages = {14},
location = {Vancouver, BC, Canada}
}

@article{cornelio23,
	title = {Combining data and theory for derivable scientific discovery with {AI}-{D}escartes},
	volume = {14},
	journal = {Nat. Comm.},
	author = {Cornelio, Cristina and Dash, Sanjeeb and Austel, Vernon and Josephson, Tyler R. and Goncalves, Joao and Clarkson, Kenneth L. and Megiddo, Nimrod and  El Khadir, Bachir and Horesh, Lior},
    pages = {1777},
    year = {2023},
    doi = {https://doi.org/10.1038/s41467-023-37236-y}    
}

@article{corywright24,
	title = {Evolving scientific discovery by unifying data and background knowledge with {AI} {H}ilbert},
	volume = {15},
	journal = {Nat. Comm.},
	author = {Ryan Cory-Wright and Cristina Cornelio and Sanjeeb Dash and El Khadir, Bachir and Lior Horesh},
    pages = {5922},
    year = {2024},
    doi = {https://doi.org/10.1038/s41467-024-50074-w}    
}

@article{cabanas25,
  author = 	 {Cabanas-Tirapu, Oriol and Danús, Lluís and Moro, Esteban and Sales-Pardo, Marta and Guimerà, Roger},
  journal = 	 {Nat. Comm.},
  title = 	 {Human mobility is well described by closed-form gravity-like models learned automatically from data},
  volume = {16},
  pages = {1336},
  year = {2025},
  doi = {http://dx.doi.org/10.1038/s41467-025-56495-5}
}

@Article{minotaki24,
author ="Minotaki, Maria G. and Geiger, Julian and Ruiz-Ferrando, Andrea and Sabadell-Rendón, Albert and López, Núria",
title  ="A generalized model for estimating adsorption energies of single atoms on doped carbon materials",
journal  ="J. Mater. Chem. A",
year  ="2024",
volume  ="12",
issue  ="18",
pages  ="11049-11061",
publisher  ="The Royal Society of Chemistry",
doi = {https://doi.org/10.1039/D3TA05898K}
}

@article{jog24,
title = {Hybrid analytical surrogate-based process optimization via Bayesian symbolic regression},
journal = {Comput. Chem. Eng.},
volume = {182},
pages = {108563},
year = {2024},
issn = {0098-1354},
doi = {https://doi.org/10.1016/j.compchemeng.2023.108563},
author = {Sachin Jog and Daniel Vázquez and Lucas F. Santos and José A. Caballero and Gonzalo Guillén-Gosálbez}
}

@article{artime21,
    author = {Artime, Oriol and De Domenico, Manlio},
    journal = {Nat. Comm.},
    volume = {12},
    pages = {2478},
    title = {Percolation on feature-enriched interconnected systems},
    year = {2021},
    doi = {https://doi.org/10.1038/s41467-021-22721-z}
}

@article{guimera26,
    author = {Guimerà, Roger and Sales-Pardo, Marta},
    title = {Bayesian symbolic regression: automated equation discovery from a physicist’s perspective},
    journal = {Philos. Trans. R. Soc. A},
    volume = {384},
    number = {2317},
    pages = {20250089},
    year = {2026},
    month = {04},
    issn = {1364-503X},
    doi = {10.1098/rsta.2025.0089}
}

\end{document}